\documentclass[letterpaper, 10 pt, conference]{ieeeconf}
\IEEEoverridecommandlockouts

\usepackage{graphicx}

\usepackage{amsmath, amsfonts, amssymb}
\usepackage{url}
\usepackage{algorithm}
\usepackage{algpseudocode}
\usepackage{subcaption}
\usepackage{arydshln}
\usepackage{multirow}
\usepackage{adjustbox}

\usepackage{mathtools}

\newcommand{\STAB}[1]{\begin{tabular}{@{}c@{}}#1\end{tabular}}

\usepackage{todonotes}

\usepackage{amsthm}

\makeatletter
\let\NAT@parse\undefined
\makeatother
\usepackage{hyperref}
\hypersetup{
   colorlinks=true,
    linkcolor= blue,
    allcolors=blue,
    citecolor = blue,
    filecolor=black,
    urlcolor=blue,
    }

\title{Modeling the Lane-Change Reactions to Merging Vehicles for Highway On-Ramp Simulations}

\author{ Dustin~Holley,
Jovin~D'sa,
Hossein~Nourkhiz Mahjoub,
Gibran~Ali,
Tyler Naes\\
Ehsan~Moradi-Pari,
Pawan Sai Kallepalli
\thanks{D. Holley and P. Kallepalli are with the Global Center for Automotive Performance Simulation (email: dholley@gcaps.net)}
\thanks{J. D'sa, H. N. Mahjoub, T. Naes and E. Moradi-Pari are with Honda Research Institute, USA Inc. (email: jovin{\_}dsa@honda-ri.com)}
\thanks{G. Ali is with the Division of Data and Analytics, Virginia Tech Transportation Institute (email: gali@vtti.vt.edu)}
}

\begin{document}

\maketitle

\begin{abstract}
Enhancing simulation environments to replicate real-world driver behavior is essential for developing Autonomous Vehicle technology. While some previous works have studied the yielding reaction of lag vehicles in response to a merging car at highway on-ramps, the possible lane-change reaction of the lag car has not been widely studied. In this work we aim to improve the simulation of the highway merge scenario by including the lane-change reaction in addition to yielding behavior of main-lane lag vehicles, and we evaluate two different models for their ability to capture this reactive lane-change behavior.
To tune the payoff functions of these models, a novel naturalistic dataset was collected on U.S. highways that provided several hours of merge-specific data to learn the lane change behavior of U.S. drivers. To make sure that we are collecting a representative set of different U.S. highway geometries in our data, we surveyed 50,000 U.S. highway on-ramps and then selected eight representative sites. The data were collected using roadside-mounted lidar sensors to capture various merge driver interactions. 
The models were demonstrated to be configurable for both keep-straight and lane-change behavior.
The models were finally integrated into a high-fidelity simulation environment and confirmed to have adequate computation time efficiency for use in large-scale simulations to support autonomous vehicle development. 

\end{abstract}

\section{Introduction}

\PARstart{M}{erging} at highway on-ramps has been a topic of interest in the transportation research community due to its significant impact on traffic flow, safety, and efficiency \cite{margiotta2011agency}, \cite{6338779}. Improving the efficiency of the highway on-ramp merging scenario can have a big impact on alleviating traffic congestion in cities \cite{marczak2013merging}, \cite{rios2016survey}, \cite{sun2014modeling}. Additionally, many drivers consider merging onto the highway to be a challenging and stressful scenario, especially during rush hour traffic, due to the need to negotiate with other drivers in both a short distance and time-sensitive manner \cite{zgonnikov2020should}. 

Autonomous vehicles (AVs) and Advanced Driver Assistance Systems
(ADAS) have been an active research topic in academia and the automotive industry, and these technologies have the potential to make driving safer, smoother, and more efficient. Given the importance of the highway on-ramp merging scenario, much research has been devoted to developing algorithms for AVs that assist the driver to merge onto the highway.
Having a high fidelity simulation environment is essential to benchmark and validate these different algorithms and all of these studies also emphasized the importance of accurate modeling of the main lane traffic participants' behavior to improve the performance of the designed merging algorithms.

Recent studies have examined the merge interactions between merging and non-merging vehicles in an attempt to provide insights into the important factors in developing interaction-aware behavior models.  An empirical analysis of merging behavior was provided in \cite{daamen2010empirical}. In \cite{yue2022effects}, the effect of an AV merging on main lane traffic during the merging scenario was discussed. In addition, Wang et al. \cite{wang2021social} investigated the impact of social preferences on merging vehicle decisions. Together, these studies have begun to illustrate the complex nature of merge interactions. 

In relation to modeling the reaction to a merging car for simulation improvement, several researchers have studied the yielding or longitudinal reaction to a merging car \cite{choudhury2009modeling}. The Merge Reactive Intelligent Driver Model (MR-IDM) proposed in \cite{holley2023mr} models the yielding dynamics of the traffic vehicles and was tuned using a large real-world merging dataset. When it comes to the lane change interactions at merge on-ramps, most studies have focused on the lane changing of the merging subject vehicle \cite{kang2017game} \cite{chen2023game}. However, the potential lane change reaction of a main lane lag vehicle to a merging car, which is a valid reaction, hasn't been widely studied. AV and ADAS technology for highway on-ramp merging can benefit from improved simulation environments that can capture and represent different driver interactions, including the lane change behavior of traffic vehicles. We try to address this gap in this work. To the best of our knowledge, commercially available simulators do not include a merge-interaction-specific simulator. While several models for general discretionary lane change (DLC) have been proposed and validated \cite{MOBIL}, the lane change reaction behavior at highway on-ramps is more difficult to model.
Models that show good performance for describing DLC at highways don't always generalize well enough when studying the reaction to a merging car.  
In this work we will utilize the MR-IDM model proposed in \cite{holley2023mr} for the longitudinal model of the traffic vehicle and will evaluate two different lane change decision making models that will run alongside this model to cover a wider range of traffic vehicle reactions to be used in simulation.  

\subsection{Related Work}
The literature includes extensive efforts trying to model lane-changing behavior in different driving situations. In this section, we provide an overview of different modeling methodologies in the literature and briefly discuss their relevance to our goals in this work.

Lane change modeling literature can be divided into two main categories of lane change decision making and lane change inference. While lane change decision making models attempt to recreate and mimic the lane change agent’s decision making before a lane change, 
lane change inference models focus on predicting and/or detecting an agent’s lane change maneuver prior to or during the maneuver execution.

Since our target in this work is developing an appropriate model that can create the proper lane change behaviors for the main lane traffic agents who are reacting to the merging agent, the first modeling category, i.e., \textit{decision making} models, is more relevant. 
Lane change \emph{decision making} modeling ranges from simplistic rule-based approaches, such as Gipps-like, mechanistic models as in \cite{Gipps}, MOBIL \cite{MOBIL}, and LMRS \cite{LMRS}, game theoretical models as in \cite{HLGT,BRGTD,BRGTM}, Markov process models such as \cite{Toledo}, and artificial intelligence (AI) methods such as fuzzy logic \cite{fuzzy} and artificial neural networks \cite{NeuralDriverAgents}. 

An important requirement for our modeling framework is being efficient enough to be integrated in high-fidelity simulators. Neural network and AI methodologies are known for their high data requirements \cite{LCpdReview} and computational cost, which may not be suitable for simulated traffic. They are also prone to over-fitting \cite{CompModels}. These methodologies, along with machine learning approaches like generative and discriminative models, also have a reduced interpretability for simulation results analysis. Looking at less complex modeling approaches, simple rule-based models can lack generalizability to situations not considered during development. While not as interpretable as Gipps-type models, game-theoretic models follow an understandable logic, require fewer computational resources than AI methods, and are a popular choice for modeling stochastic, human-like behavior. Therefore, game-theoretic methods are the most favorable modeling category for our goals in this work.

\subsection{Contributions of the Paper}
Vehicle simulation software tools, such as IPG-CarMaker \cite{carmaker2021reference}, CarSim \cite{manual2002mechanical}, or VISSIM \cite{ptv2018ptv}, are widely used for testing autonomous systems against a variety of on-road situations. Within these simulation environments, non-ego traffic agents are often given a simple, rule-based model such as the Wiedemann models \cite{Wiedemann74,durrani2016calibrating} or the Intelligent Driver Model \cite{IDM}, which were not specifically designed or modified for merge negotiations. Meanwhile, merge-specific models generally restrict the merging actor's lag vehicle behavior (hereafter referred to as $Lag_0$) to longitudinal axis, only simulating changes in acceleration. The real-time computational performance of these models is also rarely presented.

This paper offers two main contributions to merge traffic analysis and modeling. First, we collected the HOMER (Highway On-ramp MERging) dataset which is a novel highway merging dataset. A large survey of U.S. on-ramps was performed to identify a subset of eight merge ramps for data collection. The data collected from these eight sites are representative of a large percentage of national merging traffic. Second, this work adapts microscopic discretionary lane change (DLC) models to operate in merge negotiations alongside the longitudinal MR-IDM, allowing the traffic actor $Lag_0$ to react to the merging actor with both longitudinal and lateral behavior. Strengths and weaknesses of the models, including their ability to replicate real-world lane-change decisions, are presented. These models are also shown to operate effectively in a real-time simulation environment at a high update rate.

\subsection{Organization of the Paper}
The structure of the paper is as follows. In Section \ref{sec:DataCollection}, we introduce the datasets used and related data processing necessary for this work, which includes the introduction of the novel HOMER dataset. In Section \ref{sec:ProblemSetup}, we introduce our modeling framework and explain the two primary models we evaluated in this study, including the formulation of the proposed mBRGT-D model. We then present the results in Section \ref{sec:results} and discuss the performance of the proposed model. We provide concluding remarks and future directions in Section  \ref{sec:conclusion}.

\section{Data Collection and Processing}\label{sec:DataCollection}

To evaluate the validity of performing a lane change, $Lag_0$ must be able to observe and interact with actors in the prospective target lane and with actors in its current lane and the merge lane.  It is thus important that data used to train and validate a lane change model are able to sufficiently capture all relevant actors for an acceptable period of time. We found that the quality of actor trajectories in datasets collected by relatively stationary sensors, such as a camera-mounted drone, or a camera and lidar sensor mounted on a high pole, were better suited than those taken from the ego-vehicle perspective, as they were able to better avoid traffic occlusions. The exiD \cite{exiDdataset} and the Honda HOMER datasets, which have on-ramp merging specific data, were selected. 

\subsection{The HOMER Dataset}

\begin{figure}
    \centering
        \includegraphics[width=0.98\linewidth]{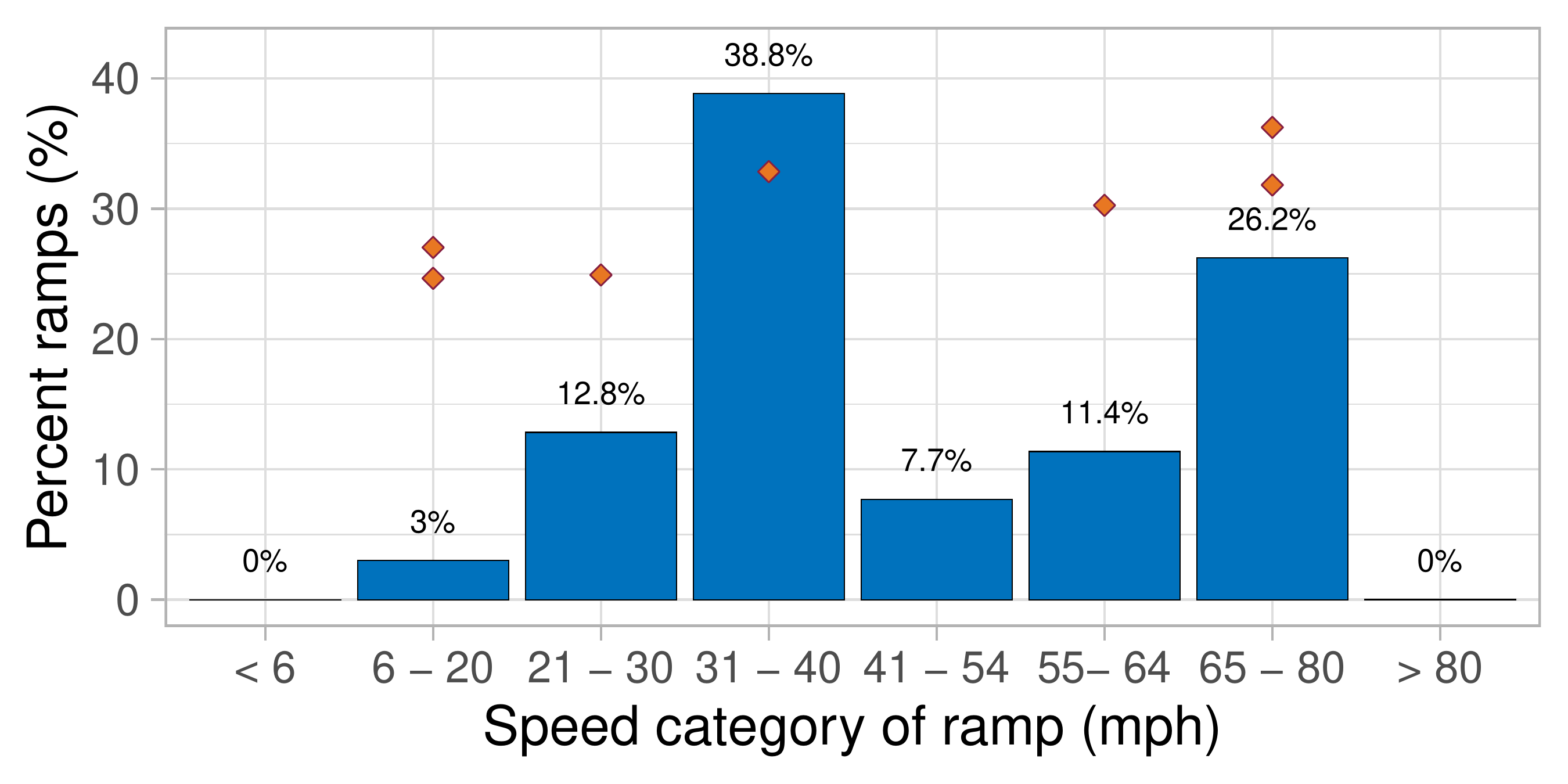}  
        \label{fig_site_selection_rmp_speed_cat}
    \includegraphics[width=0.99\linewidth]{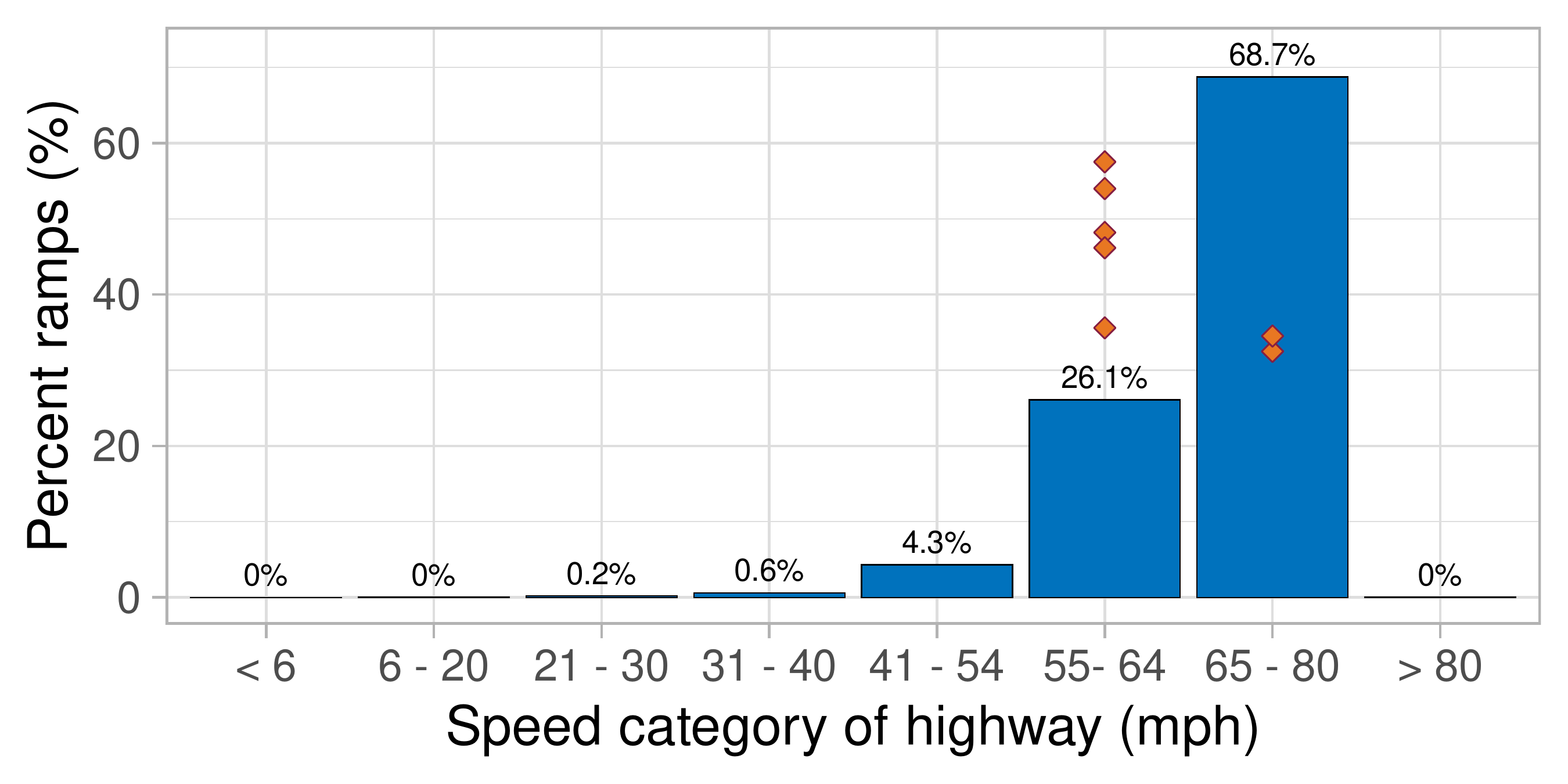}  
        \label{fig_site_selection_hw_speed_cat}
    \includegraphics[width=0.99\linewidth]{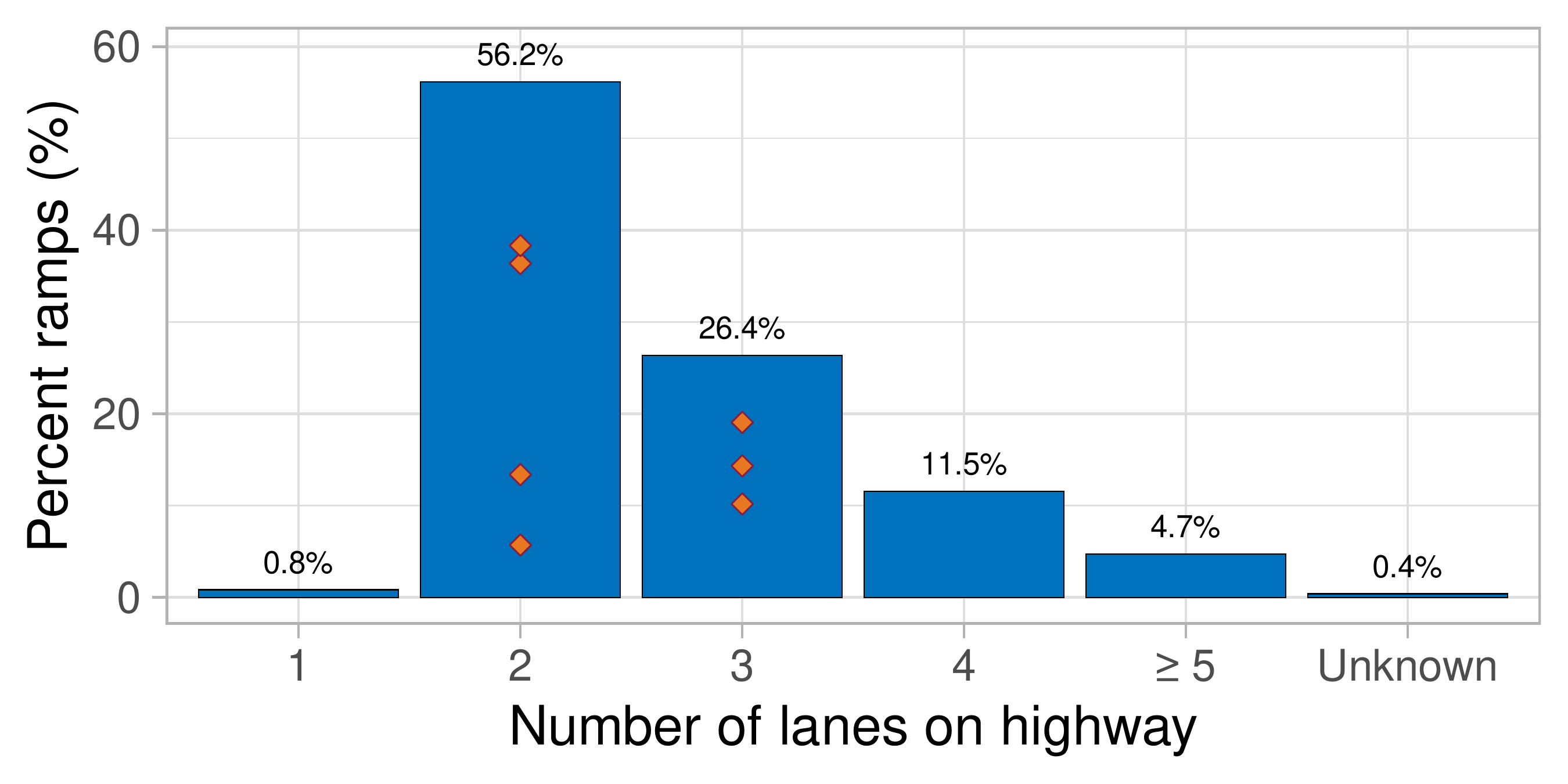}  
        \label{fig_site_selection_num_lanes}
        \includegraphics[width=0.99\linewidth]{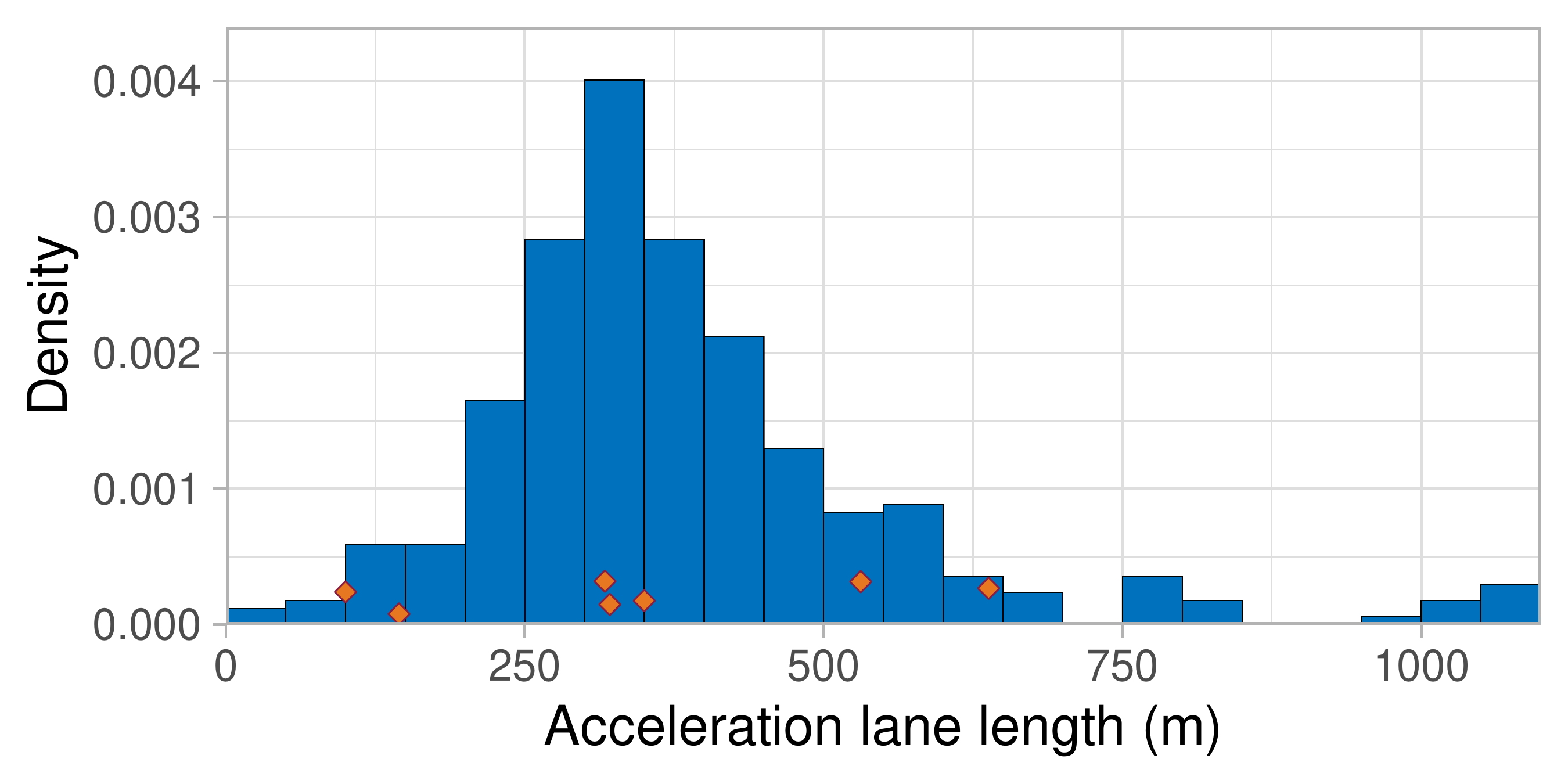}  
        \label{fig_site_selection_lane_lngth}
        \includegraphics[width=0.99\linewidth]{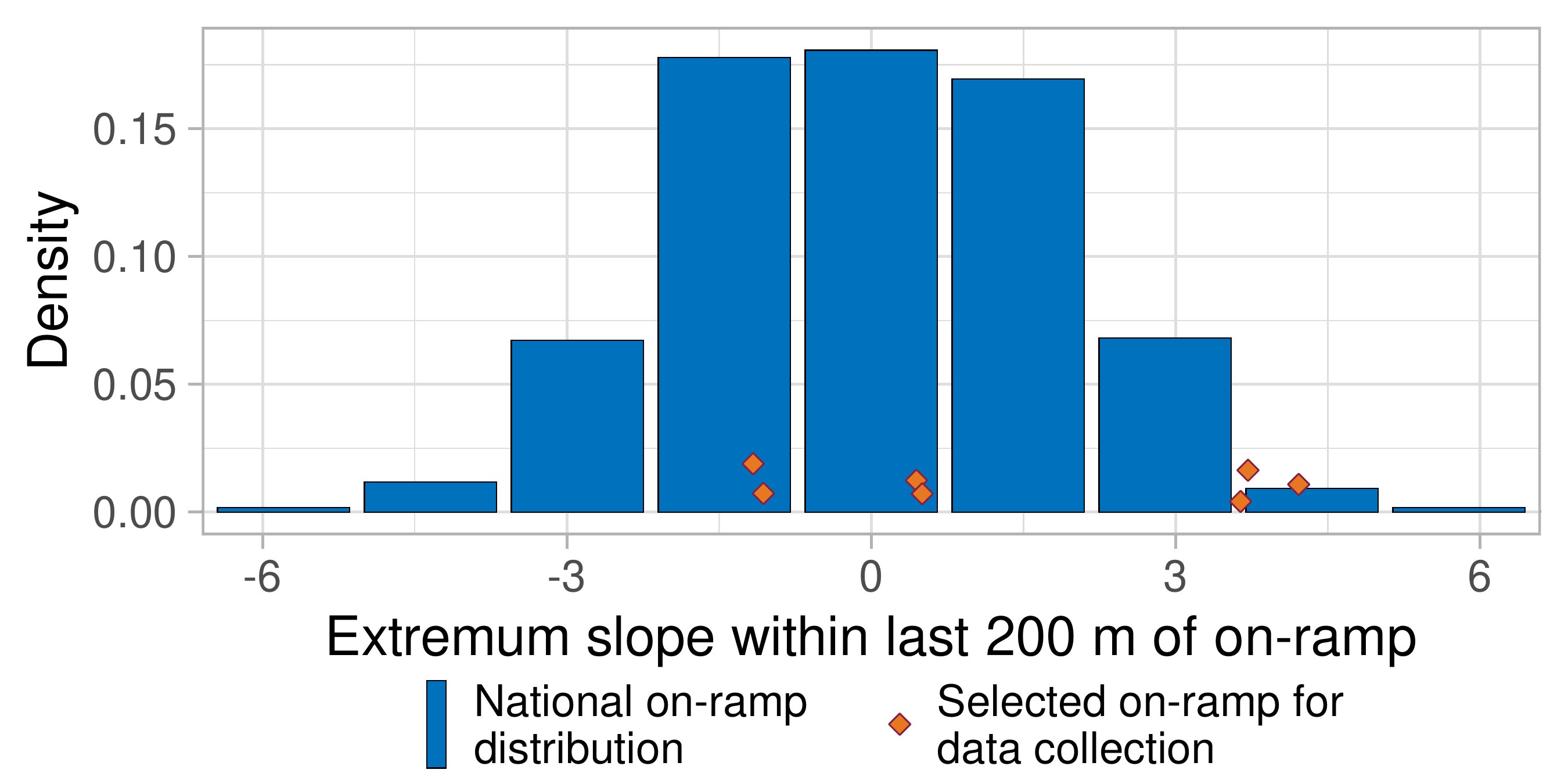}  
        \label{fig_site_selection_slope}
    \caption{The distribution of national on-ramps and data collection sites selected to maximize range of speed category of on-ramp, speed category of highway, number of lanes on highway, acceleration lane length, and the on-ramp slope.}
    \label{fig_site_selection_stats}
\end{figure}

The HOMER dataset was created to capture and understand traffic dynamics around on-ramp merging areas. To do so successfully, the dataset needed to fulfill three primary requirements. First, it was necessary to obtain the traffic negotiation recordings from a variety of merging areas at on-ramps, representing a significant portion of typical driving scenarios across the United States. Secondly, it needed to successfully capture various types of traffic negotiations during merging through a robust sensor suite. Finally, it needed to capture many scenarios that could be used to understand and simulate the norms and outliers experienced during merging. 

To find ideal data collection sites, an extensive survey of over 50,000 U.S. on-ramps was conducted. This survey determined the distribution of key factors that can affect the traffic dynamics around merging areas, such as speed differential between the ramp and the highway, number of traffic lanes, acceleration lane length, and the extremum slope of the on-ramp before merging. The team then selected the data collection sites iteratively to cover a large range of the distribution. The blue bars in Fig. \ref{fig_site_selection_stats} illustrate the distribution of these key metrics for the overall U.S. dataset. The horizontal position of the orange diamonds represents the values of these metrics for the sites selected for data collection. This figure clearly shows that for each of the metrics, 80-90\% of the range is covered by the data collection sites. The data collection was carried out in eight sites, with three based in Southwest Virginia, two based in Northern Virginia near Washington, DC, two based in Michigan, and one based in Ohio. 

\begin{figure}[!h]
  \centering
   \includegraphics[width=0.9\columnwidth,height=2.0cm]{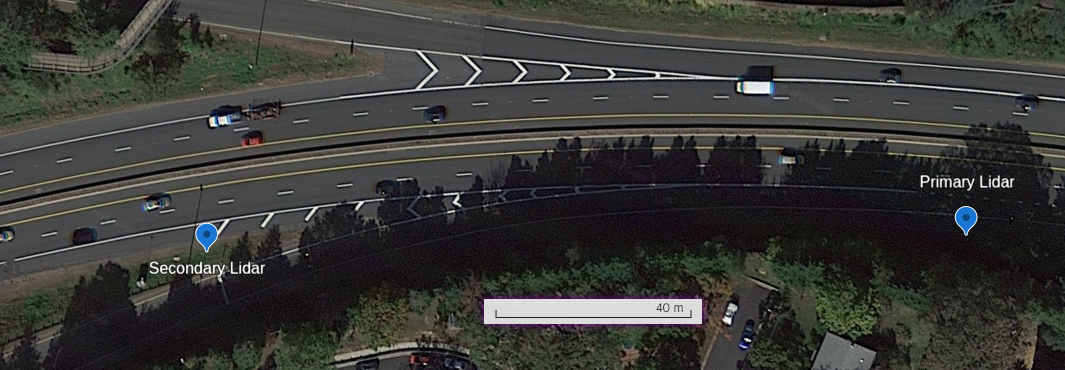}\vspace{0.1cm}
    \includegraphics[width=0.9\columnwidth, height=4.5cm]{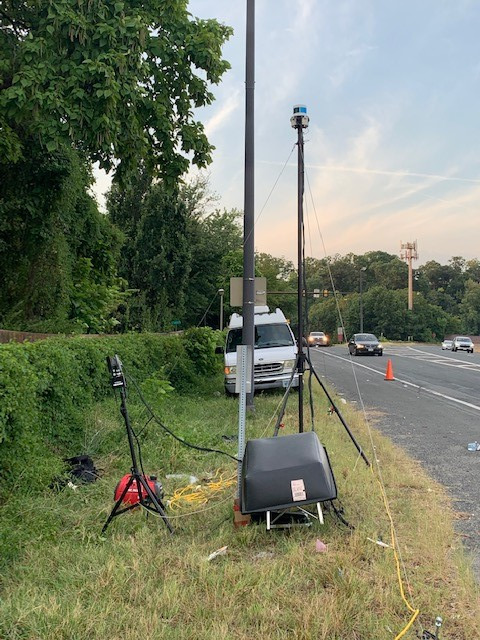}\vspace{0.1cm}
    \includegraphics[width=0.9\columnwidth, height=4.5cm]{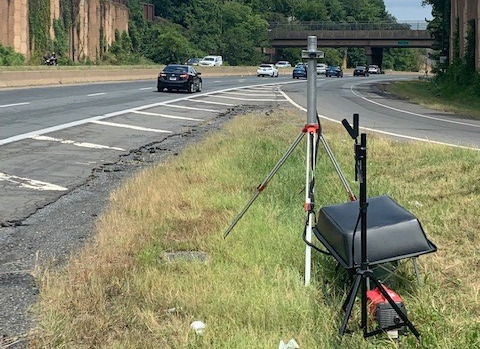}
  \caption{The data collection setup (top) included two lidar sensors placed on the side of the road: a 128-beam primary lidar placed near the merging area (middle) and a 32-beam secondary lidar placed further down the road (bottom). Each lidar sensor was mounted on a base station mast, which included a camera platform, power supply, 
  cellular WiFi router, data acquisition server, and monitor.}
  \label{fig:datacollection_setup}
\end{figure}

Several data collection methods were evaluated to capture the traffic dynamics around merging areas. It was determined that using two roadside lidars would be the most effective way to capture the traffic interactions accurately over an extended period of time. Fig. \ref{fig:datacollection_setup} shows three images. The top image shows the site plan for one of the data collection sites located in Northern Virginia near Washington DC. The figure also shows that two lidars were used to capture the traffic movement, with a primary 128-beam lidar (middle) placed further ahead of the merging area and a secondary 32-beam lidar (bottom) placed near the start of the merge. The placement of the two lidars ensured that the traffic dynamics are captured from the pre-merge highway, on-ramp and merging area, and post-merge highway regions. The data were collected over several hours at each of the locations and stored in local hard drives connected to the data acquisition system.  

\begin{figure*}[t]
    \centering
    \includegraphics[width = 0.90\textwidth]{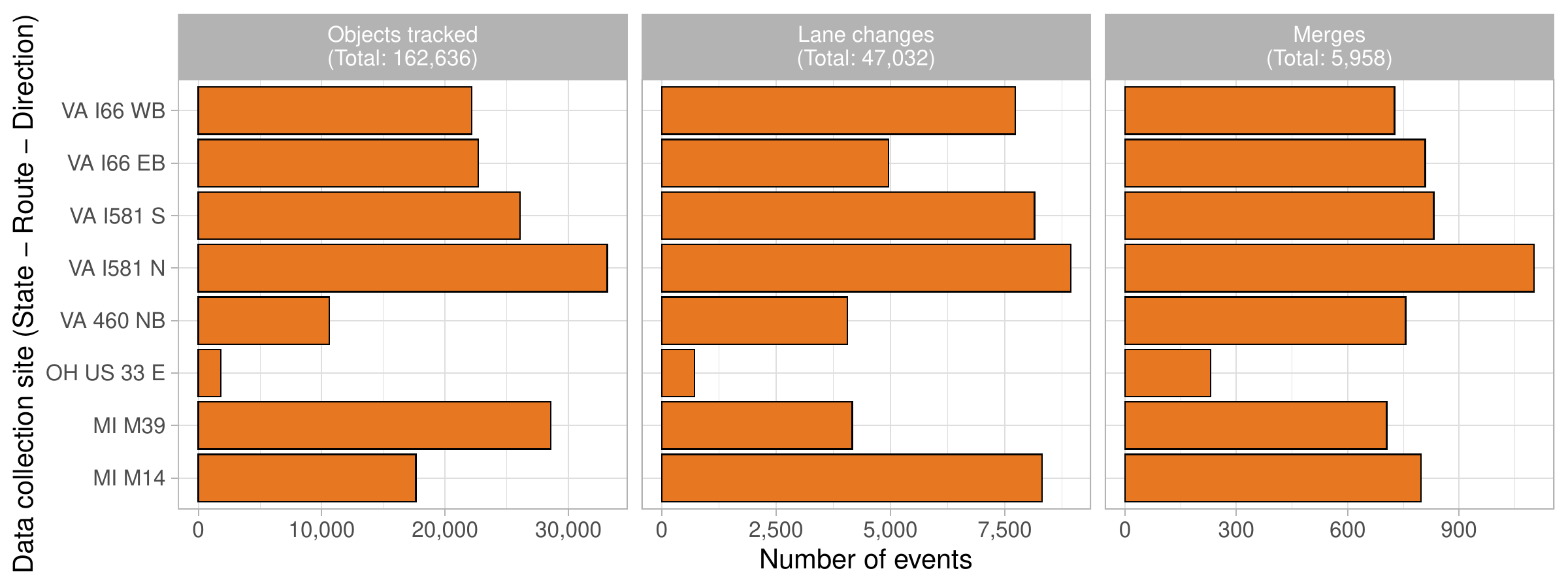}
    \caption{Summary of the HOMER dataset collected across eight US sites in terms of total number of objects tracked, number of lane changes, and number of merges processed.}
    \label{fig:HOMER_data_Summary}
\end{figure*}

Once the data collection was complete, the data were transferred to compute clusters where various data processing steps were performed. These steps included data cleaning, lidar data processing, data augmentation with roadway attributes, and summarizing the data.  Figure \ref{fig:HOMER_data_Summary} illustrates the overall data collection metrics in terms number of objects tracked, number of lane changes detected, and number of merges collected for each site.

\subsection{exiD Dataset}
The exiD dataset \cite{exiDdataset} consists of naturalistic road user trajectories focusing on exits and entries of highways in Germany. This dataset contains data from seven different locations and provides accurate positional information along with map lanelet information that can be used to parse the road geometry.

\section{Problem Setup}\label{sec:ProblemSetup}

\subsection{Data Extraction and Preparation}\label{sec:DataExtrAndPrep}

\begin{figure}
    \vspace{6pt}
    \centering
    \includegraphics[width=0.48 \textwidth,keepaspectratio]{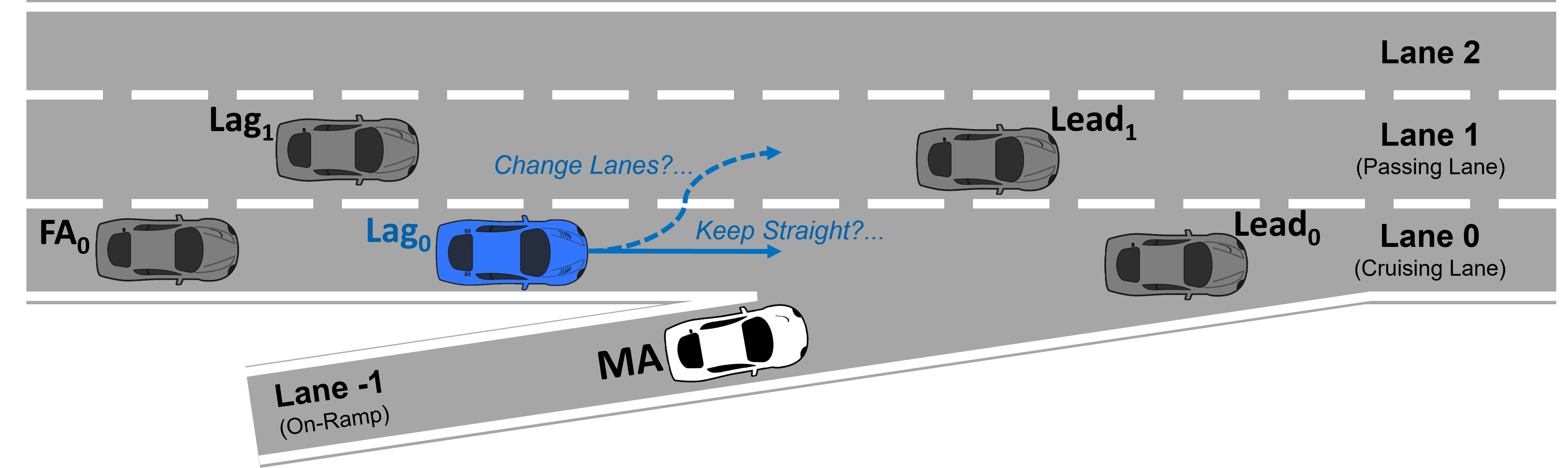}
    \caption{Layout of relevant actors and their designations.}
    \label{fig:actorlayout}
\end{figure}

A key requirement of the data collection sites was that they must also conform to the target scenario, which, in our case, is a merge negotiation between $Lag_0$ and Merging Actor ($MA$), while $MA$ is traveling on an ending merge ramp, with the presence of an additional passing lane (Lane 1) to allow for lane change behavior of $Lag_0$. Focusing on the datasets with stationary sensors, HOMER contained seven valid collection sites, exiD contained five valid sites. 
Of the remaining sites, one site's on-ramp was situated to the left of the traffic lanes (HOMER's Ann Arbor site). The rest of the sites contained ramps that merged into traffic from the right.
For the remaining sites, the relevant actors and lanes were subsequently identified (see Fig. \ref{fig:actorlayout}).

A total of $\sim$26.7 hours of HOMER data and $\sim$13.7 hours of exiD data were processed. From these data, all actors that could be labeled "$Lag_0$" were found. From those, only actors that met the following criteria were kept:
\begin{itemize}
    \item The actor must not occupy Lane -1 (On-Ramp lane) at any point. This criterion removes actors that may be refusing to change lanes due to a desire to merge out of traffic in the case of a cloverleaf merge area and also removes actors that are merging into traffic themselves.
    \item The actor must qualify as a $Lag_0$ (a traffic actor that is the lag to a merging actor) and be within 60 meters longitudinally of a merging vehicle for more than 5 seconds. This removes negotiation periods that are too short for valid evaluation. If $Lag_1$ is present at any point during this time, then only the duration in which $Lag_1$ is present is counted. This prevents the model from being evaluated based on decisions made prior to any knowledge of $Lag_1$'s existence.
\end{itemize}
The final tally of $Lag_0$ actors considered is as follows:
\begin{itemize}
    \item Keep straight: 3,331 actors (97.7\%) in HOMER, 647 actors (92.2\%) in exiD
    \item Change lanes: 80 actors (2.3\%) in HOMER, 55 actors (7.8\%) in exiD
\end{itemize}
In total, 3.3\% of valid $Lag_0$ instances changed lanes.
These findings are in agreement with \cite{MCEGT}'s conclusion that the frequency of $Lag_0$ selecting the lane-changing strategy is relatively low, as seen in naturalistic datasets such as the NGSIM \cite{NGSIM} dataset. 

Two lane change models from the literature were chosen for implementation and comparison: MOBIL \cite{MOBIL} and Wang et al.'s game theoretic DLC model \cite{BRGTD}, hereafter referred to as BRGT-D (Bounded Rationality Game Theory - DLC). MOBIL was chosen as the baseline lane change model for this project due to its popularity as a DLC model, interpretability, and reliance on an acceleration input that can be easily generated using the MR-IDM. BRGT-D was chosen due to its interpretable structure, tunability, and incorporation of lead-vehicle influences that could be used to incorporate the effect of a merging actor through input from the MR-IDM.

As noted in Section \ref{sec:DataExtrAndPrep}, only 3.3\% of the extracted merge events contained a lane-changing $Lag_0$. This imbalance in the data inhibits the proper tuning of a single model configuration to handle both keep straight and lane change behavior modalities. Therefore, for this analysis, two configurations were targeted for both models to target the two known behaviors (keep straight and lane change).

\subsection{General Modeling Procedure}

Both models were implemented with the following general procedure. At each time step, if a $Lag_0$ actor is found:
\begin{enumerate}
    \item Given any relevant actors in Lanes 0 and 1, calculate the projected current accelerations (and Deceleration Rates to Avoid a Crash [DRACs] for BRGT-D) associated with the two $Lag_0$ cases:
    \begin{itemize}
        \item $Lag_0$ is in Lane 0 (keep straight).
        \item $Lag_0$ is in Lane 1 (change lanes).
    \end{itemize}
 and the two $Lag_1$ cases (for BRGT-D only):
    \begin{itemize}
        \item $Lag_1$ does not decelerate (do not yield).
        \item $Lag_1$ decelerates (yield).
    \end{itemize}
All acceleration values are calculated using MR-IDM with the following selected parameter set: $s_0 = 2$, $T = 1.6$, $a = 0.73$, $b = 1.67$, $v_0 = 33.33$ (in concurrence with \cite{BRGTD}), and $\zeta = 1$. The MR-IDM calculations consider both $MA$ and $Lead_0$.
    \item Run the lane change model using the calculated values.
    \item If the model chooses to change lanes, assign the actor a lateral trajectory as a function of time.
\end{enumerate}

\subsection{MOBIL Model}
The MOBIL \cite{MOBIL} equations with asymmetric lane bias were implemented with little modification. Acceleration values for non-existent actors (for example, when $Lag_1$ does not exist) were set to 0 m/s$^2$.
Negative values were included in the bounds for MOBIL's politeness parameter ($p$) to allow for more aggressive behavior, as discussed in \cite{pCalib}, modifying the bounds from (0,1) to (-3,3).
Since $Lag_0$ always initially occupies the cruising lane, lane bias was assumed to always favor the current lane. Thus, the sum of $\Delta{a_{th}}$ and $\Delta{a_{bias}}$ was combined into a single $\Delta{a_{th}}$ term to reduce the number of model parameters.

\subsection{Bounded Rationality Game Theory Model (DLC)}

BRGT-D \cite{BRGTD} uses the projected accelerations ($a$) of $Lag_0$ and $Lag_1$ and their projected DRAC. DRAC is defined as:
\begin{equation}
    DRAC_{F}^{L}=\frac{(v_{L}-v_{F})^2}{(s_{L}-s_{F}-l_{F})}, v_{F}>v_{L}
\label{eq:BRGTD1}
\end{equation}
where $v$ is speed, $s$ is longitudinal position, $l$ is vehicle length, $L$ is the leader, and $F$ is the follower.

The payoff functions for $Lag_0$ and $Lag_1$ are as follows:
\begin{equation}
    \begin{aligned}
    \pi_{Lag_0} = &\omega_1a_{Lag_0,ij}+\omega_2DRAC_{Lag_0,ij}^{Lead_1}\\
    &+\omega_3DRAC_{Lag_1,ij}^{Lag_0}+\omega_4DRAC_{Lag_0,ij}^{Lead_0}
    \end{aligned}
\label{eq:BRGTD2}
\end{equation}
\begin{equation}
    \pi_{Lag_1} = \mu_1a_{Lag_1,ij}+\mu_2DRAC_{Lag_1,ij}^{Lead_1}+\mu_3DRAC_{Lag_1,ij}^{Lag_0}
\label{eq:BRGTD3}
\end{equation}
where $\omega_n$ and $\mu_n$ are coefficients for calibration, $i$ is $Lag_0$'s action, and $j$ is $Lag_1$'s action.

The probability that an actor will choose the highest payoff is then determined based on the bounded rationality $\beta$ of the actor through a quantal response equilibrium (QRE):
\begin{equation}
\left\{ 
  \begin{array}{ c l }
    \displaystyle P_{Lag_0}(s_i) =  \frac{e^{E\pi_{Lag_0}(s_i,F_j)/ \beta }}{\displaystyle\sum_{s_i'\epsilon S}e^{E\pi_{Lag_0}(s_i',F_j)/ \beta}} \\
     \displaystyle P_{Lag_1}(f_i) =  \frac{e^{E\pi_{Lag_1}(S_i,f_j)/ \beta }}{\displaystyle\sum_{f_i'\epsilon F}e^{E\pi_{Lag_1}(S_i,f_j')/ \beta}}
  \end{array}
\right.
\label{eq:BRGTD4}
\end{equation}
where $E\pi_{n}$ is actor $n$'s expected payoff, given the other actor's actions.

$\beta$ can be set to decrease over the negotiation to indicate increasing rationality:
\begin{equation}
    \beta(t) = \beta + (\alpha-\beta)e^{-\delta(t-1)}
\label{eq:BRGTD5}
\end{equation}
where
\begin{equation}
    \alpha = \sum_{i=0}^{P}m_ik^i
\label{eq:BRGTD6}
\end{equation}
and $P=4$, $m_0$-$m_P$ are parameters to tune, and $k$ is the traffic density of the target lane.

The BRGT-D model was first implemented as given in \cite{BRGTD}, with the proposed parameter values. As with the MOBIL implementation, acceleration and DRAC values for non-existent actors defaulted to 0 m/s$^2$.
Mixed Nash equilibrium was calculated using the Lemke-Howson \cite{LemkeHowson} algorithm to identify $Lag_1$'s expected action. $Lag_0$'s corresponding payoffs were then used to find the probabilities that $Lag_0$ would change lanes or keep straight.

Several modifications were made to adapt the BRGT-D to our use case, with the subsequent model referred to as mBRGT-D (merge-adapted BRGT-D). The acceleration value calculations were performed using the MR-IDM instead of the original IDM in order to incorporate the merging actor's influence on $Lag_0$ as well as the leader. In \cite{BRGTD}, the focus was on optimizing the parameters involved in the $\beta$ calculation. However, $\beta$ only influences the outcome by adjusting the probability of choosing the highest payoff between 50\% and 100\% (adjusting "rationality"). For our merging data, the model was found to tend toward choosing a lane change in most cases, so the payoff calculation needed to be adjusted. Thus, the payoff parameters $\omega_n$ and $\mu_n$ were chosen as the optimization targets.
The model's bounded rationality equations (involving $\beta$) were omitted from the optimization parameter list. In this way, we were able to focus on optimizing the payoff functions to target reactionary lane changes appropriately without stochastic variability in the results, since we were more interested in achieving the correct ultimate decision than in adjusting the probability of performing the decision. This also helped reduce the number of considered parameters during optimization.

To provide further control over the model's lane-changing decision, an additional term was added to $Lag_0$'s payoff function, shown in Equation \ref{eq:BRGTD7}. This term introduces a lane bias, as used in the MOBIL equations, to penalize lane changes to the passing lane.
\begin{equation}
    \pi_{Lag_0,new} = \pi_{Lag_0} + \omega_5\lambda_1\lambda_2
\label{eq:BRGTD7}
\end{equation}
where
\begin{align}
    \nonumber \lambda_1 &= 
        \left\{ 
          \begin{array}{ c l }
            -1,& \: \textrm{if action = change lanes} \\
             1,& \: \textrm{if action = keep straight} \\
          \end{array}
        \right.
        \nonumber \\
    \nonumber \lambda_2 &= 
        \left\{ 
          \begin{array}{ c l }
            -1,& \: \textrm{if targeting the cruising lane} \\
             1,& \: \textrm{if targeting the passing lane} \\
          \end{array}
        \right.
        \nonumber
\label{eq:BRGTD8}
\end{align}
and $\omega_5$ is a new parameter.

\subsection{Parameter Optimization}
Two optimization approaches were considered for evaluation. The first approach targeted overall prediction success rate using a single parameter value set for both the MOBIL and mBRGT-D models. For the second approach, two parameter sets were targeted, corresponding to the two defined lateral behaviors: a lane-changing set and a keep-straight set. The dataset was divided into training and validation sets. The training dataset contained 70\% of both the HOMER and exiD datasets, also containing 70\% of both keep-straight and lane-change cases. The validation set contained the other 30\%. MATLAB's \emph{MultiStart} \cite{MultiStart} optimizer was used for optimization.

For the first approach, due to the low lane-change rate in the raw data, using the overall prediction success rate ($r_{overall}$) as the sole input to the cost function would result in a parameter set that avoids lane changes. To place more emphasis on the lane-change cases, a weighted sum of the individual keep straight ($r_{KS}$) and lane change ($r_{LC}$) prediction success rates were included in the cost, with similar weights given to the two rates.
Also, to ensure that the optimizer does not focus on a local minimum wherein $r_{LC}$ is high but $r_{KS}$ is low, or wherein $r_{LC}$ and $r_{KS}$ are equivalent but the overall prediction rate is low, $r_{overall}$ is included as a third value. To prevent $r_{overall}$ and $r_{KS}$ from dominating, as their magnitudes have a direct correlation, their weights were dropped below that of $r_{LC}$. The cost function to minimize was formulated as follows:
\begin{equation}
    \begin{aligned}
    cost = 0.4(100\%-r_{LC}) &+ 0.3(100\%-r_{KS})\\
    &+ 0.3(100\%-r_{overall})
    \end{aligned}
    \label{eq:opt1}
\end{equation}

The second approach's resultant sets were targeted to enable control of simulated lag vehicle lateral behavior. The goal was to maximize the successful prediction rate of the model on all cases corresponding to the target behavior:
\begin{equation}
cost = 
\left\{ 
  \begin{array}{ c l }
    100\%-r_{KS},& \: \textrm{if targeting keep-straight cases} \\
     100\%-r_{LC},& \: \textrm{if targeting lane-change cases}
  \end{array}
\right.
\label{eq:opt2}
\end{equation}

A third approach, wherein a distribution of parameter value sets was generated by optimizing for each case individually, was considered. However, this was not pursued due to the lack of a sufficient number of lane-change cases.

\section{Results and Discussion}\label{sec:results}
The overall optimization results are shown in 
Table \ref{tab:BRGTopt3}
. Optimization was unable to achieve reasonable results for both lane-change and keep-straight prediction success rates, although overall prediction success rate could be achieved if the keep-straight prediction rate was high. 
It was found that lower bounds for mBRGT-D resulted in high lane-change success rates but low keep-straight success, so the upper bounds for this model were increased. This may be due in part to the generally small DRAC values and, to a lesser extent, small acceleration values. 
Since the goal was not to develop a lane-change prediction model, but rather to produce a simulation model that could emulate lane-change decisions and be tuned for desired behavior, evaluation shifted to the second, behavior-specific optimization approach.

 The behavior-specific optimization results are shown in 
 Tables \ref{tab:BRGTopt4} and \ref{tab:BRGTopt5}
 .
High prediction success could be reasonably achieved with both models. While mBRGT-D maintained high success rates for both lateral behaviors, MOBIL's performance declined for lane changes. 
MOBIL's politeness factor ($p$) was tuned to a negative (aggressive) value to improve the lane-change success rate. mBRGT-D's higher adaptability when compared to MOBIL can be attributed in part to the model's higher number of considered influencing factors and parameters.

\begin{table}[h]
\caption{Results for Overall optimization. Both training (T) and validation (V) results are shown.}
\centering
\renewcommand{\arraystretch}{1.0}
\begin{adjustbox}{width=245pt,center}

\begin{tabular}{|c|c c c | c c c|}
\hline
 & Param & Bounds & Value & $r_{KS}$ & {$r_{LC}$} & {$r_{overall}$} \\
\hline
\hline
\multirow{4}{*}{\STAB{\rotatebox[origin=c]{90}{     MOBIL   }}}
 & $b_{safe}$ &  ( 0, 4 ) & 3.26 & & & \\
 & $\Delta{a_{th}}$   &  ( 0, 4 ) & 1.35 & T=91.8\% & T=13.7\% & T=89.2\% \\
 & $p$        &  (-3, 3 ) & 1.91 & V=91.6\% & V=20.0\% & V=89.3\% \\
 & $\Delta{a_{bias}}$ &  - & 0.00 & & & \\
 \hline
\hline 
\multirow{7}{*}{\STAB{\rotatebox[origin=c]{90}{     mBRGT-D \hspace{0.1cm}   }}}

 &  $\omega_1$ &  ( 0, $1e3$ ) & 905.79 & & & \\
 &  $\omega_2$ &  ( 0, $1e4$ ) & 1621.82 & & & \\
 &  $\omega_3$ &  ( 0, $1e4$ ) & 7948.31 & & &\\
 &  $\omega_4$ &  ( 0, $1e4$ ) & 5004.72 & T=64.2\%  & T=77.9\% & T=64.7\% \\
 &  $\omega_5$ &  ( 0, $1e4$ ) & 2955.34 & V=65.5\%  & V=80.0\% & V=65.9\% \\
 &  $\mu_1$ &  ( 0, $1e3$ ) & 312.72 & & &\\
 &  $\mu_2$ &  ( 0, $1e4$ ) & 8419.29 & & &\\
 &  $\mu_3$ &  ( 0, $1e4$ ) & 4951.77 & & &\\
 \hline
\end{tabular}
\end{adjustbox}
\label{tab:BRGTopt3}
\end{table}

 \begin{table}[h]
\caption{Results for Keep-Straight optimization. Both training (T) and validation (V) results are shown.}
    \centering
    \renewcommand{\arraystretch}{1.25}
\begin{tabular}{| c| c c c | c |}
 \hline
 & Param\hspace{0.2cm} & Bounds\hspace{0.2cm} & Value & $r_{KS}$ \\
 \hline
\multirow{3}{*}{\STAB{\rotatebox[origin=c]{90}{     MOBIL \hspace{0.1cm}   }}}
 & $b_{safe}$ &  ( 0, 4 ) & 0.22 & \\
 & $\Delta{a_{th}}$   &  ( 0, 4 ) & 3.72 & T=99.2\%  \\
 & $p$        &  (-3, 3 ) & 0.46 & V=99.2\%  \\
 & $\Delta{a_{bias}}$ &  - & 0.00 &\\
 \hline
 \hline
\multirow{8}{*}{\STAB{\rotatebox[origin=c]{90}{  mBRGT-D  \hspace{0.25cm}}}}
& $\omega_1$ &  ( 0, $1e3$ ) & 11.90 &\\
& $\omega_2$ &  ( 0, $1e4$ ) & 5107.72 &\\
& $\omega_3$ &  ( 0, $1e4$ ) & 9786.81 &\\
& $\omega_4$ &  ( 0, $1e4$ ) & 225.13 & T=97.3\% \\
& $\omega_5$ &  ( 0, $1e4$ ) & 4852.29 & V=97.2\% \\
& $\mu_1$ &  ( 0, $1e3$ ) & 662.01 &\\
& $\mu_2$ &  ( 0, $1e4$ ) & 1362.93 &\\
& $\mu_3$ &  ( 0, $1e4$ ) & 9391.42 &\\
 \hline
 \hline
\multirow{8}{*}{\STAB{\rotatebox[origin=c]{90}{  mBRGT-D without $\omega_5$ }}}
 & $\omega_1$ &  ( 0, $1e3$ ) & 11.90 & \\
 & $\omega_2$ &  ( 0, $1e4$ ) & 5107.72 & \\
 & $\omega_3$ &  ( 0, $1e4$ ) & 9786.81 &\\
 & $\omega_4$ &  ( 0, $1e4$ ) & 225.13 & T=53.6\%  \\
 & $\omega_5$ &  ( 0, $1e4$ ) & - & V=51.4\%  \\
 & $\mu_1$ &  ( 0, $1e3$ ) & 662.01 &\\
 & $\mu_2$ &  ( 0, $1e4$ ) & 1362.93 &\\
 & $\mu_3$ &  ( 0, $1e4$ ) & 9391.42 &\\
 \hline
\end{tabular}
\label{tab:BRGTopt4}
\end{table}

 \begin{table}[h]
     \vspace{5pt}
\caption{Results for Lane Change optimization. Both training (T) and validation (V) results are shown.}
    \centering

     \renewcommand{\arraystretch}{1.25}

\begin{tabular}{ |c| c c c | c | }
 \hline
 & Param\hspace{0.2cm} & Bounds\hspace{0.1cm} & Value & $r_{LC}$  \\
 \hline
 \hline
\multirow{4}{*}{\STAB{\rotatebox[origin=c]{90}{  MOBIL}}}
 & $b_{safe}$ &  ( 0, 4 ) & 3.36 & \\
 & $\Delta{a_{th}}$   &  ( 0, 4 ) & 0.24 & T=91.6\%  \\
 & $p$        &  (-3, 3 ) & -2.18 & V=85.0\%  \\
 & $\Delta{a_{bias}}$ &  - & 0.00 &\\
 \hline
 \hline
 \multirow{8}{*}{\STAB{\rotatebox[origin=c]{90}{  mBRGT-D }}}
 & $\omega_1$ &  ( 0,  10 ) & 5.06 &\\
 & $\omega_2$ &  ( 0, 10 ) & 3.90 & \\
 & $\omega_3$ &  ( 0, 10 ) & 0.38 & \\
 & $\omega_4$ &  ( 0, 10 ) & 1.71 & T=95.8\%   \\
 & $\omega_5$ &  ( 0, 10 ) & 0.62 & V=95.0\%  \\
 & $\mu_1$ &  ( 0,  10 ) & 4.71 & \\
 & $\mu_2$ &  ( 0, 10 ) & 4.24 & \\
 & $\mu_3$ &  ( 0, 10 ) & 6.95 & \\
 \hline
\end{tabular}
\label{tab:BRGTopt5}
\end{table}

Note that, for both MOBIL and mBRGT-D, targeting prediction success for \emph{lane-change} cases resulted in a significant decrease in the prediction success rate for \emph{keep-straight} cases, and vice versa. There could be several potential reasons for this inverse relationship. First, the model inputs may be inadequate to represent the entire space of influencing situational factors on lane changes. Other factors may provide a better predictor of lane-change decisions. Second, there may be significant enough heterogeneity in drivers' decision making, given similar circumstances, that a single model configuration could not obtain an accurate overall prediction success rate.
The latter conclusion would seem to support the argument that accurate lane change inference may depend on the inclusion of lateral state data (technically \emph{detection} as opposed to \emph{prediction}) instead of solely relying on situational data.

The addition of the $\omega_5$ parameter to the mBRGT-D model provided a marked improvement in keep-straight prediction performance. 
Table \ref{tab:BRGTopt4} also 
shows the optimization performance of the mBRGT-D model without $\omega_5$. Due to the bias toward lane-changing behavior, optimization cannot achieve a high prediction success rate. Note that optimization was not able to find a more suitable parameter set than the one found for mBRGT-D with $\omega_5$.

For the random distribution of a model's resultant parameter sets among simulated traffic, the sample weights of the two parameter sets can correspond to the keep straight (96.7\%) and lane change (3.3\%) rates found in the data. This restricts simulated traffic to approximate a real-world lane-change rate under appropriate conditions, using situational context to perform the decision-making. The keep-straight configuration can perform lane-changes in safety-critical situations, and vice versa. This offers a more realistic improvement over the assignment of blind macroscopic lane-change rates to simulated traffic. The lane-change sample rate can also be increased to simulate more accommodating traffic or decreased for more aggressive, unyielding traffic.

\subsection{High-Fidelity Simulation}

\begin{figure}
    \centering
    \vspace{5pt}
    \includegraphics[width=0.2\textwidth,keepaspectratio]{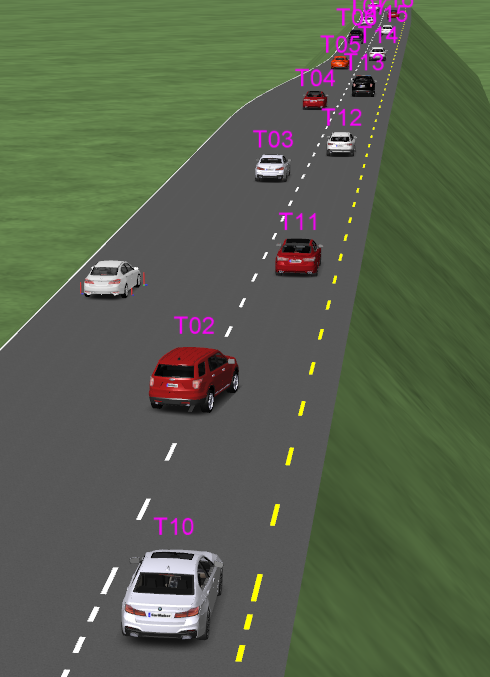}
    \caption{Snapshot of simulation in CarMaker. Actor \emph{T02} is performing a lane change in response to the merging vehicle.}
    \label{fig:carmakerlc}
\end{figure}
One of the motivations of this work was to use these proposed models in a high-fidelity simulation environment to simulate real-world merge interactions. IPG-CarMaker was used to simulate the high-fidelity vehicle dynamics of a target car along with the road environment, while MATLAB-Simulink was used to simulate the traffic behavior models. The parameters for the lane change models are sampled from the optimized values, and the actors are divided between Lanes 0 and 1.

This setup was able to successfully simulate a highway merge scene involving 17 cars, each with their own traffic behavior model and underlying dynamics at 50 Hz. This evaluation was performed on a Windows desktop with a $64$-bit OS, Intel(R) Xeon(R) W-$2123$CPU @$3.6$ GHz  processor, and $16$ GB RAM. In the evaluation, the simulation setup with either model was able to achieve real-time performance, with the isolated performance of the MOBIL and mBRGT-D model code capable of reaching an average of $6.6$x and $5.6$x real-time performance, respectively. This shows that the computational load of the game-theoretic mBRGT-D is on par with the lighter MOBIL.

\section{Conclusion}\label{sec:conclusion}
The work in this paper adapts the work of previous DLC decision making models for highway traffic actors, expanding on previous work to include forced merge scenarios for highway on-ramps. Parameters for our modified mBRGT-D and the baseline MOBIL model were optimized using over 40 hours of real-world data of traffic actors at highway on-ramps. 
From our evaluation it was found that it was difficult to find a single set of model parameters that could achieve reasonable results for both lane-change and keep-straight prediction success rates, although overall prediction success rate could be achieved if the keep-straight prediction rate was high. However, when targeting distributed parameters for each individual behavior, we were able to obtain better optimization results. The subsequent parameter sets provide direct control over actor behavior during simulation testing.
Our mBRGT-D model was better able to replicate the lane-change reaction rate as seen in the naturalistic data with a success rate of 95\% compared to the baseline MOBIL model's 85.0\% success rate. The models were then confirmed to efficiently run in a high-fidelity simulation environment, thereby achieving the primary goal of creating a simulation environment that can generate merge reactive behaviors of lane changing or yielding (if keep straight behavior is predicted) when interacting with a merging car.

Other than the use of acceleration as an input to the lane change models, lateral and longitudinal control are largely decoupled in this work. Future work can be devoted to integrating these models into a single decision making model to produce a more robust and dynamic overall traffic model. Another future area of research is the further analysis of the collected data to identify additional variables that have influence on the lane-change decision which we may not have considered, which can be used to improve model performance. This research can also be used to evaluate how much of the lane-changing decision process is dependent on the identified metrics and how much is reliant on the heterogeneous characteristics of the evaluated drivers.

\section{Acknowledgement}

The authors would like to thank Travis Terry, Takayasu Kumano, Behdad Chalaki, Kevin Kefauver, Shane McLaughlin, and Samer Rajab for their contribution to this project through its many phases, including planning, data collection, and general discussions.

\bibliographystyle{IEEEtran}
\bibliography{references}

\end{document}